\documentclass[runningheads]{llncs}

\usepackage{wrapfig}
\usepackage[scaled=0.93,proportional]{zlmtt}

\usepackage[sfdefault,condensed]{roboto}  

\usepackage[T1]{fontenc}
\usepackage{graphicx}
\usepackage{hyperref}
\usepackage{color}

\usepackage{amsmath}
\DeclareMathOperator\arctanh{arctanh}
\usepackage{multirow}
\usepackage{rotating}
\usepackage[x11names]{xcolor}
\usepackage{colortbl}
\usepackage{algorithm}

\definecolor{lightgray}{RGB}{240, 240, 240}

\definecolor{paleyellow}{HTML}{FFE3C0}
\definecolor{dullblue}{RGB}{100,100,200}
\long\def\comment[#1]#2{\par\noindent
  \ifx\void#1\else{\marginpar{\raggedright\fontsize{6pt}{6pt}\selectfont #1}}\fi
  \colorbox{paleyellow}{%
    \parbox[t]{\columnwidth}{\setlength{\parskip}{1ex plus 0.2ex minus 0.2ex}#2}}}

\usepackage{mathtools}

\usepackage{multicol}

\usepackage{pseudo}
\pseudoset{
ct-left={{~\small//~}},
ct-right=\relax,
indent-length=1em,
label={\small\arabic*},
kw,
}

\newcommand{\littleparagraph}[1]{\medskip\par\noindent\emph{#1}}

\definecolor{DodgerUniformBlue}{rgb}{0.0,0.353,0.612}
\newcommand{\define}[1]{\emph{\textcolor{DodgerUniformBlue}{#1}}}

\newcommand{\RED}[1]{\textcolor{red}{#1}}
\newcommand{\BLUE}[1]{\textcolor{blue}{#1}}
\newcommand{\YELLOW}[1]{\textcolor{yellow}{#1}}
\newcommand{\PURPLE}[1]{\textcolor{purple}{#1}}

\usepackage{etoolbox}
\AtBeginEnvironment{thebibliography}{\scriptsize}

\begin{document}
\title{Highlighting Case Studies in LLM Literature Review of Interdisciplinary System Science \thanks{This paper is published in AI2024: Advances in Artificial Intellgience, DOI: 10.1007/978-981-96-0348-0\_3}}
\titlerunning{Case Studies in LLM Literature Review}
%
\author{Lachlan McGinness\inst{1,2}\orcidID{0000-0002-3231-4827},  
Peter Baumgartner\inst{2,1}\orcidID{0000-0002-6559-9654} \\ Esther Onyango\inst{3}\orcidID{0000-0003-0116-4923} \and Zelalem Lema\inst{4}\orcidID{0000-0001-9275-8828}} 
\authorrunning{L. McGinness, P. Baumgartner et al.}
%
\institute{Australian National University, 
\email{first.last@anu.edu.au}\\
\and
Data61|CSIRO, 
\email{first.last@data61.csiro.au}
\and
CSIRO, Agriculture and Food, \email{Esther.Onyango@csiro.au}
\and
CSIRO, Agriculture and Food, \email{Zelalem.Moti@csiro.au}
}
\maketitle              
\begin{abstract}
Large Language Models (LLMs) were used to assist four Commonwealth Scientific and Industrial Research Organisation (CSIRO) researchers to perform systematic literature reviews (SLR). We evaluate the performance of LLMs for SLR tasks in these case studies. In each, we explore the impact of changing parameters on the accuracy of LLM responses. The LLM was tasked with extracting evidence from chosen academic papers to answer specific research questions. We evaluate the models' performance in faithfully reproducing quotes from the literature and subject experts were asked to assess the model performance in answering the research questions.
We developed a semantic text highlighting tool to facilitate expert review of LLM responses.  

We found that state of the art LLMs were able to reproduce quotes from texts with greater than 95\% accuracy and answer research questions with an accuracy of approximately 83\%. We use two methods to determine the correctness of LLM responses; expert review and the cosine similarity of transformer embeddings of LLM and expert answers. The correlation between these methods ranged from 0.48 to 0.77, providing evidence that the latter is a valid metric for measuring semantic similarity.

\end{abstract}
\section{Introduction}

The scientific community is currently full of hype and hope for the use of Artificial Intelligence (AI) to accelerate research  \cite{Nature2024Why}. Messeri and Crockett claim that scientists are too trusting and lack awareness of the biases and errors of Large Language Models (LLMs) \cite{Messeri2024Artificial}. They present `AI as Oracle' as a vision of the future where LLMs overcome the problem of too much literature to digest by efficiently searching and summarising information \cite{Messeri2024Artificial}. Many research groups 
are optimising and improving LLM tools for literature review
\cite{Brody2021Scite,Kung2023Elicit,Aguilera2024Accelerating} including the CSIRO's
(Commonwealth Scientific and Industrial Research Organisation's) Science Digital program \cite{CSIRO2024Trust}.

Many tools exist to enhance or automate literature review, including LitLLM \cite{Agarwal2024LitLLM}, Scite \cite{Brody2021Scite}, Elicit \cite{Kung2023Elicit} and Scopus AI \cite{Aguilera2024Accelerating}. The techniques of these tools remain undisclosed as commercial secrets. However they all appear to use a combination of the same strategies: calling LLMs through APIs, prompt engineering (in-context learning), Retrieval Augmented Generation (RAG) and fine tuning \cite{Bolanos2024Artificial}. The lack of transparency about these tools make it difficult to determine and accelerate best practice in AI systematic review methods.  

Systematic Literature Review (SLR) was developed in the field of Evidence-Based Medicine as a method of reducing bias by sticking to strict protocols \cite{Bolanos2024Artificial}. It has since been adopted in many disciplines including social science, education and environmental science \cite{Bolanos2024Artificial}.
SLR tasks include planning, searching, screening, extraction and synthesis\cite{Bolanos2024Artificial}. In the extraction phase, desired information is extracted from a set of selected studies. 
Few studies have attempted to objectively measure the capability of LLMs for SLR \cite{Spillias2023Human,Ye2024Hybrid,DeSilva2024AIInsights,Sami2024System,Shaib2024Summarizing} and even fewer for the extraction and screening phases. Previous studies found that LLMs are unreliable SLR tools as they `hallucinate' references that do not exist \cite{Smith2024Reviews}.

In this paper we evaluate the performance of GPT-3.5 Turbo, and GPT-4 Turbo, one of the best models currently available, on SLR tasks. We will refer to these models as GPT3 and GPT4 respectively. 
We present four case studies where LLMs were used to assist CSIRO interdisciplinary systems science researchers, 
including the last two authors of this paper, in different stages of SLR. In each we systematically explore the impact of changing a parameter on the accuracy of LLM responses.

Although automatically checking LLM responses is highly desirable, currently there are no tools that can perfectly check the correctness or semantic similarity of texts. An important and tedious step of using LLMs in SLR is verifying their responses. To make this task easier we contribute a \emph{highlighting} algorithm, in analogy to highlighting with a text marker on paper. It aims to provide a human reader with visual clues to quickly scan generated text. The algorithm is driven by a small set of user-supplied keywords provided by a domain expert. We describe the algorithm and report on experiments and experiences with application to our case studies.

The rest of this paper is structured as follows.
Section~\ref{sec:methodology} summarizes our methodology for application to the case
studies chosen. This includes statistical evaluation methods and LLM techniques.
Section~\ref{sec:semantic-text-highlighting} introduces an algorithm for explainable 
text relevance in terms of semantic similarity, and communicating it through text highlighting.
Section~\ref{sec:results} reports on experimental results, 
and Section~\ref{sec:discussion} discusses these results and Section~\ref{sec:conclusion} summarises the conclusions.

\section{Methodology}
\label{sec:methodology}
We explore four case studies with interdisciplinary system scientists. All four studies used Microsoft Azure endpoints to call either GPT3 and GPT4. The first case study focuses on the health impacts of agri-food transitions. 
In the second case study, the researcher had performed an SLR extracting the enablers and constraints from twelve papers on coordinated responses to crises. They were interested to know if they had missed any key points when completing the SLR. 
The third case study verifies an SLR involving sixty papers on a sustainable transitions SLR task \cite{Moallemi2024Early}.
The final case study focuses on screening papers for an SLR on the use of generative AI for marking student responses to exam questions.
The studies investigate the impact on overall performance when using different models, asking LLMs to provide evidence for their answers, and splitting tasks into several calls. 

An example research question from the first case study is ``What are the health outcomes of an agri-food transition''? The LLM was tasked with answering the question and finding evidence from an academic paper to support its answer. 
Evidence would be a quote from the paper such as ``An overabundance of food supply alone has been identified as a key cause of the obesity epidemic'' and the LLM answer could be ``Obesity''.

\subsection{Statistical and Evaluation Techniques}
In this section we outline the statistical methods used to analyse the results.
In cases where multiple similar data are available these results are summarised using mean ($\mu_x$) and population standard deviation ($\sigma_x$) defined as usual. 

In our analysis we use two methods to determine the correctness of LLM responses; expert review and automated similarity metrics of LLM and expert answers. 
The fist automated similarity metric is SpaCy Semantic Similarity \cite{SpacySimilarity}. This method compares two strings the average embedding vector of the tokens in each of the strings is calculated and the cosine similarity is taken. The second method is the cosine similarity between transformer-based embeddings \cite{Ormerod2021Predicting,Turton2021Deriving,Laskar2020Contextualised}, which we will refer to as `\textit{transformer similarity}'. Transformer based embeddings have the ability to take order of words into account.

To compare these two metrics we use Pearson correlation coefficient shown in Equation \ref{eq:Corr}.
\begin{equation}
    \text{Correlation} = r = \frac{\sum_{i=1}^n (\mu_x-x_i)(\mu_y-y_i)}{\sigma_x \sigma_y}
    \label{eq:Corr}
\end{equation}

Uncertainty in correlation values was calculated using the Fisher transformation of correlation \cite{Fisher1915Frequency} with a 95\% confidence interval ($Z = 1.96$) using Equation \ref{eq:FisherUncertainty}. In order to make these values more comparable to standard deviation they were divided by $Z$.

\begin{equation}
    \text{Correlation Uncertainty} = \Delta r =  \frac{1}{Z} \tanh \left( \arctanh (r) \pm \frac{Z}{\sqrt{n-3}}  \right)
    \label{eq:FisherUncertainty}
\end{equation}

One of the SLR tasks involves the using an LLM to screen papers and determine their
relevance. If a paper is relevant we refer to this as positive. A true positive (TP)
result is when a model classifies a relevant paper (as determined by experts) as
relevant. A false positive (FP) occurs when an irrelevant paper is classified as
relevant. True negatives (TN) and false negatives (FN) are defined similarly. 
False positive and False Negative rates in the normal way.

\subsection{LLM Techniques}
We converted the research papers from PDF to text with the \texttt{pdftotext} utility\footnote{\url{https://www.xpdfreader.com/pdftotext-man.html}}. Instead of using standard RAG techniques, we provided the entire paper to the LLM as the context windows were large enough. The research scientists were concerned by the biases of LLMs \cite{Blodgett2020Language,Rogers2021Changing,Bender2021DangersOf}. Here we outline these and our approach to minimising them.

\littleparagraph{Selection bias:} The LLM might favour, for example, well known authors or recent papers. To avoid this type of bias, the research scientist selected the papers for the SLRs.

\littleparagraph{Inability to understand nuance:} There are many studies which note that Large Language Models are effective for general tasks but can struggle with domain specific knowledge and the nuance of specialised tasks \cite{Chang2024Survey,Meng2024Application,Li2023Nuances}. We were concerned that LLMs may lack the nuance to determine whether information relates to the study in the paper or referenced studies. To overcome this issue, researchers removed specific pages based on the task that the LLM needed to perform. For example, when asking about the geographic location of the paper, only the abstract and introduction were provided. We also addressed this issue by using in-context learning to provide the model with specific definitions provided by domain experts.

\littleparagraph{Lack of domain specific knowledge:} The domain or even subject specific vocabulary in scientific literature poses challenges for LLM-based analysis. LLMs are trained on vast corpuses of which the subject specific matter is only a small portion. In agri-food transitions and co-ordinated responses to crises research litterature terms have very specific meanings and LLMs would miss the nuances of these words. We mitigated this problem by providing the LLMs with definitions of subject specific vocabulary as part of the prompts. 

\littleparagraph{Hallucinations:} 
It is well known that LLMs can `make up' information \cite{Ji2023Survey}. To manage this, tasks were broken into two steps. First, the Large Language Model was asked to find quotes that provide evidence of the desired information. Binary (yes/no) verification of the quotes failed because of trivial errors related to unicode characters. Therefore quotes were instead verified using fuzzy text matching using \texttt{thefuzz} implementation\footnote{\url{https://github.com/seatgeek/thefuzz}} of the Levenshtein distance metric~\cite{levenshtein1966binary}.  The LLM was then given the quotes and asked for a final answer. 

\section{Semantic Text Highlighting}
\label{sec:semantic-text-highlighting}
In this section we introduce a method for \emph{semantic text highlighting}, or \define{highlighting} for short. The idea of semantic highlighting is not new and was originally formulated for information retrieval~\cite{hussam_semantic_1998}.
In our context we apply highlighting to LLM retrieved evidence. Because the amount of retrieved evidence can still be overwhelming, further support is needed to aid a human reviewer. This is where highlighting comes in.

Like with a text marker on paper, highlighting does not need to be perfect to be useful. However, the highlighted words should be \emph{semantically related} to a given specific SLR research question. Our method requires a set of keywords representing the research question. Highlighting then boils down to determining which words in a given sentence are semantically related to the keywords and conveying the findings in a useful way.
For that, we rely on readily available information-theoretic similarity measures and a carefully curated corpus of English, which requires no training and is explainable.

Our method works as follows. The SLR researcher provides sets of keywords comprising of \emph{entities} $E$, \emph{relations} $R$ and \emph{properties} $P$ as nouns, (possibly transitive) verbs and adjectives or adverbs, respectively.
For example, in our first SLR case study on the topic \emph{influences of agri-food transitions on health outcomes} the following were used:
\begin{description}
\item[Entities $E$:] {\sffamily health, disease, outcome, food, lifestyle}
\item[Relations $R$:] {\sffamily explain, affect, improve, stimulate}
\item[Properties $P$:] {\sffamily environmental}
\end{description}
A word $w$ in a given text is highlighted if it is deemed related to a keyword
according to the procedure \pr{Similarity}(w, C, t) in Algorithm~\ref{alg:simlilarity}, where $C = E \cup R \cup P$.
The given text is first parsed with the part-of-speech (POS) parser SpaCy~\cite{Spacy}. It assigns a grammatic role $t$ to every word
$w$ for determining the best suited subset of $C$  for similarity. The algorithm computes two similarity scores between $w$ and that subset in the range $[0,1]$. One of them is \emph{word vector similarity}, as readily provided by spaCy,
and the other is \emph{Wu-Palmer similarity}~\cite{wu_verb_1994,shenoy_new_2012} based on hypernym-reachability in WordNet~\cite{fellbaum_wordnet_1998}. 

\begin{algorithm}[h]
\caption{Similarity}
\fontsize{8.5}{9}\selectfont
\label{alg:simlilarity}
\begin{pseudo*}
  \pr{Similarity}(w, C, t) \\
  Input: \textnormal{$w$ word, $C$ keywords in canonical form (lemmas), $t$ type of $w$ (noun, verb, adjective ...)}\\
  Output: \textnormal{Similarity score for $w$}\\
  $\pr{P-Weight} \gets 0.95$\qquad $\pr{RF-Weight} \gets 0.95$ \qquad   $\pr{WUP-Threshold} \gets 0.8$\qquad  $\pr{VEC-Threshold} \gets 0.95$\\
  $\id{best}_{\tn{wup}} \leftarrow \max_{c\in C} \pr{WUP-X}(w, c, t)$ or else $0.0$ \\
  $\id{best}_{\tn{vec}} \leftarrow \max_{c\in C} \pr{VEC}(w, c)$ or else $0.0$ \\
  if $\id{best}_{\tn{wup}} \ge \pr{WUP-Threshold}$ and $\id{best}_{\tn{wup}} \ge \id{best}_{\tn{vec}}$
  then return $\id{best}_{\tn{wup}}$ \\
  elif $\id{best}_{\tn{vec}} \ge \pr{VEC-Threshold}$
  then return $\id{best}_{\tn{vec}}$ \\
  else return $0.0$\\
  \\
  \pr{WUP-X}(w, c, t) \ct{Extended Wu-Palmer similarity, considers reachable words from $w$ and $c$}\\
  $S_w = \pr{Extend}(\pr{WN-Synsets}(w, t))$ \ct{\pr{WN-Synsets} returns synonyms of $w$}\\
  $S_c = \pr{Extend}(\pr{WN-Synsets}(c, t))$ \\
  return $\max_{(s_w \stackrel{\omega_w}{\longrightarrow} r_w, s_c
    \stackrel{\omega_c}{\longrightarrow} r_c) \in S_w \times S_c} \omega_w \cdot \omega_c \cdot \pr{WN-WUP}(r_w,
  r_c, t)$ \ct{Wu-Palmer from WordNet}\\
  \\
\pr{VEC}(w, c) \ct{Vector similarity}\\
  if $w = c$ then return $1.0$ \\
  elif \tn{both $w$ and $c$ have vector embeddings} then\\+
  return \tn{cosine-similarity of the embeddings of $w$ and of $c$}\\-
  else return $0.0$\\
  \\
  \pr{Extend}(S, t) \\
  Input: \textnormal{$S$ a set of WordNet synsets}\\
  Output: \textnormal{Weighted extension of $S$ by pertainyms and derivationally related forms}\\
  $R \leftarrow \emptyset$ \ct{Result relation}\\
  for $s \in S$ do\\+
  $R \leftarrow R \cup \{ s \stackrel{1.0}{\longrightarrow} s\}$ \ct{$R$ is reflexive}\\
    for $l \in \pr{WN-Lemmas}(s)$ do\\+
  $R \leftarrow R \cup \{ s \stackrel{\pr{P-Weight}}{\longrightarrow}
  \pr{WN-SynSet}(v) \mid v \in \pr{WN-Pertainyms}(l)\}$\\
  $R \leftarrow R \cup \{ s \stackrel{\pr{RF-Weight}}{\longrightarrow}
  \pr{WN-SynSet}(v) \mid v \in \pr{WN-RelatedForms}(l)\} $\\-- 
return $R$
\end{pseudo*}
\end{algorithm}

Using WordNet for word similarity is an established and well-researched topic \cite{manna_fuzzy_2010}. In our algorithm, the search for similar words is broad and includes (one-step) synonyms, pertainyms, and related derived forms, weighted for each category.
This was a design choice motivated by the case study in Section~\ref{sec:crisis-SLR}, where the researcher wants to ensure they did not miss any key points in their manual review. 
\pr{Similarity} takes scores higher than given thresholds to determine whether $w$ should be highlighted or not. 
We found that Wu-Palmer similarity often produces results more similar to what a human user expects from a highlighter. Hence, vector similarity acts only as a fall-back. 
Here are some examples for highlighted evidence text with the keywords above. 
\begin{enumerate}
  \itshape\small
\item
It is also likely that \RED{climate change} \YELLOW{will} \BLUE{contribute} to
\PURPLE{novel} \RED{occurrences} of \RED{disease emergence} and transmission.

\item
\RED{Foodborne illnesses} significantly \BLUE{influence} individuals \PURPLE{nutritional} \RED{status}.

\item
\RED{Changing lifestyles}, mainly due to \RED{work commitment},
\YELLOW{have} \BLUE{fuelled} the \RED{increase} in
numbers eating out and the need for \RED{convenience foods}. 

\item
  \PURPLE{Significant} \RED{changes} \YELLOW{have} \BLUE{occurred} in \RED{food systems} in the last
 decades that \YELLOW{have} \BLUE{contributed} to widen \PURPLE{such} \RED{'holes'} in the barriers from
 phase to phase: agricultural intensification and industrialization \BLUE{causing} major
 environmental deterioration, the \RED{increasing distance} \BLUE{traveled} by \RED{food} in \RED{global
 markets}, and the \RED{nutrition transition} towards \RED{diets} rich in \PURPLE{ultra} - \RED{processed
 food and animal protein} are the three cornerstones of \PURPLE{such} \RED{changes}.
\end{enumerate}
Entities (nouns, noun chunks) are colored red and relations (verbs) are colored
blue. Additional colors are used for supporting words according to their grammatical roles. Properties (adjectives) of colored entities are purple.

For each word, the algorithm can provide an explanation of why it is highlighted. These explanations are helpful for customising parameter settings; for example, we get:
\begin{itemize}
  \def\SMALL{\fontsize{8}{8}\selectfont}
  \small\itshape
  \item[2.] 
  \RED{Foodborne}
\RED{illnesses} {\SMALL\verb|(NCP(Foodborne illnesses, [SimilarTo('disease', 0.95, 'wup')]))|}
significantly \BLUE{influence} {\SMALL\verb|(SimilarTo('affect, 0.84, 'wup'))|} individuals
\RED{nutritional status}\\ {\SMALL\verb|(NCP(nutritional status, [SimilarTo('food', 0.91, 'wup')]))|}.
\end{itemize}
In these annotations, \texttt{NCP} means `NounChunkPart', and the similarity of the highlighted word(s) to keyword(s) is indicated as in
\verb|SimilarTo(keyword, similarity, kind)|, where \verb|'wup'| is Wu-Palmer similarity. 

Highlighting can be useful beyond marking up text excerpts.
In one of our case studies below we take the `highlighting rate' as a statistical measure to assess the similarity between the LLM's response and the researcher's benchmark evaluation.

\section{Results}
\label{sec:results}
\subsection{Case Study 1: Similarity Metrics in Agri-Food Transition SLR}
In this first case study we determine the accuracy of the chosen metrics and explore if there is a significant difference in performance between GPT3 and GPT4.
There were eight tasks (as shown in the first column of Table \ref{tab:Esther}) and ten papers. 
GPT3 and GPT4 were given identical instructions to answer each task in separate calls. We compared the researcher's and models' answers.

\begin{table}
\caption{Case Study 1: Average and standard deviation for similarity scores, fuzzy matching scores, average word count and expert judged model accuracy are presented for GPT3 and GPT4.}
\sffamily
\fontsize{8}{10}\selectfont
\centering
\label{tab:Esther}
\begin{tabular}{|c|c|c|c|c|c|c|c|c|}
\hline
Information & Complexity & Model & \begin{tabular}[c]{@{}c@{}}Quote Fuzzy\\ Matching\\ Score\end{tabular} & \begin{tabular}[c]{@{}c@{}}Model\\ Average\\ Word Count\end{tabular} & \begin{tabular}[c]{@{}c@{}}Expert\\ Average\\ Word Count\end{tabular} & \begin{tabular}[c]{@{}c@{}}Transformer\\ Similarity\end{tabular} & \begin{tabular}[c]{@{}c@{}}SpaCy\\ Similarity\end{tabular} & \begin{tabular}[c]{@{}c@{}}Model\\  Accuracy\end{tabular}  \\ \cline{1-9} 
 &  & GPT-3 & $97\pm5$ & $3.5$ & $4.8$ & $0.74\pm0.25$ & $0.58\pm0.16$ & $0.9$  \\ \cline{3-9} 
\multirow{-2}{*}{\begin{tabular}[c]{@{}c@{}}Global\\ Context\end{tabular}} & \multirow{-2}{*}{Low} & \cellcolor{lightgray}GPT-4 & \cellcolor{lightgray}$98\pm6$ & \cellcolor{lightgray}$5.2$ & \cellcolor{lightgray}$4.8$ & \cellcolor{lightgray}$0.84\pm0.06$ & \cellcolor{lightgray}$0.57\pm0.14$ & \cellcolor{lightgray}$1.0$  \\ \hline
 &  & GPT-3 & $97\pm5$ & $5.6$ & $1.6$ & $0.82\pm0.13$ & $0.57\pm0.16$ & $0.75$  \\ \cline{3-9} 
\multirow{-2}{*}{\begin{tabular}[c]{@{}c@{}}Associated\\ Health Focus\end{tabular}} & \multirow{-2}{*}{Low} & \cellcolor{lightgray}GPT-4 & \cellcolor{lightgray}$98\pm4$ & \cellcolor{lightgray}$7$ & \cellcolor{lightgray}$1.6$ & \cellcolor{lightgray}$0.85\pm0.05$ & \cellcolor{lightgray}$0.60\pm0.14$ & \cellcolor{lightgray}$0.94$ \\ \hline
 &  & GPT-3 & $95\pm10$ & $13.4$ & $5.4$ & $0.81\pm0.04$ & $0.66\pm0.08$ & $0.65$  \\ \cline{3-9} 
\multirow{-2}{*}{\begin{tabular}[c]{@{}c@{}}Transition\\ Pathway\end{tabular}} & \multirow{-2}{*}{Moderate} & \cellcolor{lightgray}GPT-4 & \cellcolor{lightgray}$97\pm8$ & \cellcolor{lightgray}$23$ & \cellcolor{lightgray}$5.4$ & \cellcolor{lightgray}$0.85\pm0.06$ & \cellcolor{lightgray} $0.65\pm0.21$ & \cellcolor{lightgray}$1.0$  \\ \hline
 &  & GPT-3 & $97\pm4$ & $32$ & $17$ & $0.83\pm0.04$ & $0.77\pm0.14$ & $0.5$ \\ \cline{3-9} 
\multirow{-2}{*}{\begin{tabular}[c]{@{}c@{}}Agri-food\\ Boundary\end{tabular}} & \multirow{-2}{*}{Moderate} & \cellcolor{lightgray}GPT-4 & \cellcolor{lightgray}$98\pm6$ & \cellcolor{lightgray}$50.6$ & \cellcolor{lightgray}$17$ & \cellcolor{lightgray}$0.87\pm0.03$ & \cellcolor{lightgray}$0.79\pm0.12$ & \cellcolor{lightgray}$0.85$ \\ \hline
 &  & GPT-3 & $99\pm7$ & $8.8$ & $6.5$ & $0.85\pm0.05$ & $0.59\pm0.16$ &$0.7$ \\ \cline{3-9} 
\multirow{-2}{*}{\begin{tabular}[c]{@{}c@{}}Public\\ Health Risk\end{tabular}} & \multirow{-2}{*}{Moderate} & \cellcolor{lightgray}GPT-4 & \cellcolor{lightgray} $97\pm6$ & \cellcolor{lightgray}$20.5$ & \cellcolor{lightgray}$6.5$ & \cellcolor{lightgray}$0.87\pm0.06$ & \cellcolor{lightgray}$0.74\pm0.17$ & \cellcolor{lightgray}$0.95$ \\ \hline
 &  & GPT-3 & $97\pm5$ & $31.3$ & $26.4$ & $0.83\pm0.03$ & $0.84\pm0.07$ &$0.25$ \\ \cline{3-9} 
\multirow{-2}{*}{Synergies} & \multirow{-2}{*}{High} & \cellcolor{lightgray}GPT-4 & \cellcolor{lightgray}$98\pm5$ & \cellcolor{lightgray}$58$ & \cellcolor{lightgray}$26.4$ & \cellcolor{lightgray}$0.81\pm0.05$ & \cellcolor{lightgray}$0.83\pm0.07$ & \cellcolor{lightgray}$0.1$  \\ \hline
 &  & GPT-3 & $97\pm5$ & $35$ & $18$ & $0.82\pm0.02$ & $0.81\pm0.08$ &$0.44$ \\ \cline{3-9} 
\multirow{-2}{*}{Constraints} & \multirow{-2}{*}{High} & \cellcolor{lightgray}GPT-4 & \cellcolor{lightgray} $98\pm5$ & \cellcolor{lightgray}$59$ & \cellcolor{lightgray}$18$ & \cellcolor{lightgray}$0.84\pm0.02$ & \cellcolor{lightgray}$0.83\pm0.07$ & \cellcolor{lightgray}$1.0$ \\ \hline
 &  & GPT-3 & $97\pm4$ & $28$ & $30$ & $0.89\pm0.04$ & $0.90\pm0.07$ & $0.88$ \\ \cline{3-9} 
\multirow{-2}{*}{\begin{tabular}[c]{@{}c@{}}Integrated \\ Solutions\end{tabular}} & \multirow{-2}{*}{High} & \cellcolor{lightgray}GPT-4 & \cellcolor{lightgray} $99\pm3$ & \cellcolor{lightgray} $50$& \cellcolor{lightgray} $30$ & \cellcolor{lightgray} $0.89\pm0.05$ & \cellcolor{lightgray} $0.89\pm0.07$ & \cellcolor{lightgray}$1.0$ \\ \hline
\end{tabular}
\end{table}

The models were asked to record three quotes (evidences), then give a final answer in a second call. Cases where the average fuzzy string similarity was less than $90\%$ were manually investigated. There were 25 and 38 cases where this occurred for GPT4 and GPT3 respectively, resulting in overall error rates of 2\% and 5\%. 

The semantic similarity between LLM and expert answers was calculated using \textit{transformer similarity} and SpaCy similarity. 
An example of an LLM/expert answer could be ``The Global context is Africa'' which has a word count of 5.
The correlation between the Transformer similarity score and the expert's judgement of the model answers was $0.48\pm0.09$ and there was almost no correlation ($-0.07\pm0.08$) between the SpaCy similarity and expert judgment.

\subsection{Case Study 2: Impact of Evidence on Coordinated Response to Crisis SLR}
\label{sec:crisis-SLR}
In this case study we compare LLM output based on two methods. In the first method (`evidence') the LLM first obtains quotes to support its answer to the question and then writes it's answer. In the second method (`direct') the LLM writes an answer without searching for or providing evidence. 
 
 Providing
evidence is a good way to increase the trustworthiness of LLM responses. However it will
increase the number of completion tokens and therefore cost. We test our highlighting algorithm as an automated similarity metric by calculating the correlation between the highlighted fraction of each expert answer and model answers.


Table \ref{tab:Zelalem} shows that the evidence method results in slightly lower SpaCy Semantic Score, vector embedding cosine similarity, human judged accuracy and highlighting correlation. The highlighting correlation score changes by a more significant margin, but also has a greater uncertainty than the other measures. 
\begin{table}[h]
\centering
\caption{Case Study 2: Prompt tokens, completion tokens, SpaCy similarity, transformer similarity, human judged accuracy and Highlighting Correlation are presented for comparison of direct and evidence-based conditions.}
\sffamily
\fontsize{8}{10}\selectfont
\label{tab:Zelalem}
\begin{tabular}{|l|c|c|c|c|c|c|c|}
\hline
Experimental & Prompt Tokens & Completion & SpaCy  & Transformer  & Human Judged & Highlighting \\
Condition & ($\times 10^3$) & Tokens & Similarity & Similarity & Accuracy & Correlation \\
\hline
Evidence & $20.2\pm5.7$ & $842\pm397$ & $0.85\pm0.06$ & $0.87\pm0.06$ & $69\%$ & $-0.18\pm0.24$ \\
\hline
Direct & $20.0\pm5.7$ & $213\pm64$ & $0.88\pm0.06$ & $0.90\pm0.04$ & $72\%$ & $0.13\pm0.25$ \\
\hline
\end{tabular}
\end{table}

The number of prompt tokens is not significantly changed by asking the model for evidence, but the number of completion tokens increases fourfold. As the majority of tokens are prompt tokens, it might be expected that the number of completion tokens would have a small impact on the overall cost. However the computational cost of running a transformer in this case is proportional to the number of completion tokens. A technique to more effectively reduce the cost using an LLM for literature review is grouping multiple tasks into a single call as explored in the next case study.

\subsection{Case Study 3: Impact of Combining Tasks on Sustainable Transitions SLR}
\label{sec:CombiningTasks}
In this case study, we compare conditions that we name `separate' and  `together'. For the separate condition, GPT4 is called ten times. In each call the paper is provided and the LLM is asked to find an answer and evidence for a research question. 
In the together condition, the model is given all tasks in one call. 
The results for both experimental conditions are in Table \ref{tab:Enayat}. Note that papers were classified into three groups according to their geographical scale as requested by the researcher. There were twenty papers in each category.

\begin{table}[h]
\centering
\caption{Case Study 3: Quote and final answer metrics for both conditions. The expert found significant errors in the first paper from the global scale responses for the together condition. They decided that the together method was not worth pursuing and did not evaluate the remaining together responses.}
\label{tab:Enayat}
\sffamily
\fontsize{7}{9}\selectfont
\begin{tabular}{|c|c|c|c|c|c|c|c|c|}
\hline
\multirow{3}{*}{\begin{tabular}[c]{@{}c@{}}Experimental\\ Condition\end{tabular}} & \multirow{3}{*}{\begin{tabular}[c]{@{}c@{}}Scale of\\ paper\end{tabular}} & \multirow{3}{*}{\begin{tabular}[c]{@{}c@{}}Prompt\\ Tokens\\ $\times10^5$\end{tabular}} & \multirow{3}{*}{\begin{tabular}[c]{@{}c@{}}Completion\\ Tokens\\ $\times10^3 $\end{tabular}} & \multirow{3}{*}{\begin{tabular}[c]{@{}c@{}}Fuzzy Text\\ Matching\end{tabular}} & \multirow{3}{*}{\begin{tabular}[c]{@{}c@{}}SpaCy\\ Similarity\end{tabular}} & \multirow{3}{*}{\begin{tabular}[c]{@{}c@{}}Transformer\\ Similarity\end{tabular}} & \multirow{3}{*}{\begin{tabular}[c]{@{}c@{}}Instances \\ of failing to  \\find quotes\end{tabular}} & \multirow{3}{*}{\begin{tabular}[c]{@{}c@{}}Expert\\ Judged\\ Accuracy\end{tabular}} \\
  &  &  &  &  &  &  &  &   \\
 &  &  &  &  &  &  &  &   \\ \hline
& \begin{tabular}[c]{@{}c@{}}Global\\ \hspace{0.1cm} \end{tabular} & $1.1\pm0.2$ & $1.5\pm0.2$ & $98.4\pm2.5$ & $0.84\pm0.4$ & $0.85\pm0.01$ & $0$ & $82\%$  \\ \cline{2-9} 
 Separate & \begin{tabular}[c]{@{}c@{}}International\\ /national\end{tabular} & $4.7\pm1.7$ & $1.3\pm0.1$ & $95.3\pm6.8$ & $0.79\pm0.12$ & $0.86\pm0.01$ & $5$ &  $84\%$ \\ \cline{2-9} 
 & \begin{tabular}[c]{@{}c@{}}Subnational\\ /Local\end{tabular} & $2.2\pm0.7$ & $1.4\pm0.2$ & $99.4\pm0.6$ & $0.83\pm0.03$ & $0.84\pm0.06$ & $2$ &  $75\%$ \\ \hline
& \begin{tabular}[c]{@{}c@{}}Global\\ \hspace{0.1cm} \end{tabular} & $0.12\pm0.02$ & $0.97\pm0.37$ & $99.5\pm0.4$ & $0.79\pm0.5$ & $0.83\pm0.03$ & $0$ & N/A  \\ \cline{2-9} 
Together & \begin{tabular}[c]{@{}c@{}}International\\ /national\end{tabular} & $0.48\pm0.16$ & $1.3\pm0.2$ & $93.1\pm7.2$ & $0.77\pm0.16$ & $0.84\pm0.01$ & $9$ & N/A  \\ \cline{2-9} 
 & \begin{tabular}[c]{@{}c@{}}Subnational\\ /Local\end{tabular} & $0.22\pm0.07$ & $0.8\pm0.2$ & $98.0\pm1.0$ & $0.81\pm0.03$ & $0.84\pm0.01$ & $1$ & N/A  \\ \hline
\end{tabular}
\end{table}



The researcher did not like the results of the together condition because GPT4 jumbled its responses resulting in a frame-shift. Despite the near tenfold increase in prompt tokens and cost, they preferred the separate condition. In the separate condition the model had no awareness of its answers to the other questions resulting in overlap of the ideas presented for each sub-question.

\subsection{Case Study 4: Prompt Variations for Automated Screening}
In the final case study the researcher wanted to automatically extract information from papers for the purpose of screening.  
The researcher had manually reviewed 14 papers from the arXiv and 20 from Pubmed of which 12 and 3 were relevant respectively. These were used as datasets to measure the performance of LLM information extraction. These datasets provide opposite extremes, one where the LLM needs to accept nearly all of the papers as relevant and another where it needs to reject nearly all the papers. 

This allowed for a study in simultaneously avoiding false positives and false negatives, see Table \ref{tab:PhysicsMarking}. Three prompts were used: `prompt relevant' which indicated that the paper was likely to be relevant, `prompt irrelevant', and a `neutral prompt' which makes no indication of the paper's relevance. 
\begin{table}[htbp]
\centering
\caption{Case Study 4: SpaCy and transformer similarity, expert verified accuracy, false positive and false negative rates for three different prompts on both datasets. The correlation between expert accuracy and the similarities were calculated to measure the quality of these metrics.}
\label{tab:PhysicsMarking}
\sffamily
\fontsize{7}{9}\selectfont
\begin{tabular}{|l|l|c|c|c|c|c|c|c|c}
\hline
Condition & Dataset & SpaCy & Transformer & Expert & Spacy-Expert & Transformer-Expert & False & False \\
& & Similarity & Similarity & Accuracy & Correlation & Correlation & Positive Rate & Negative Rate\\
\hline
Prompt & arXiv & $0.87\pm0.08$ & $0.94\pm0.05$ & $0.83$ & $0.64\pm0.08$ & $0.77\pm0.06$ & $1.00$ & $0.00$ \\
Relevant & \cellcolor{lightgray}PubMed & \cellcolor{lightgray}$0.70\pm0.17$ & \cellcolor{lightgray}$0.90\pm0.07$ & \cellcolor{lightgray}$0.86$ & \cellcolor{lightgray}$0.23\pm0.28$ & \cellcolor{lightgray}$0.69\pm0.21$ & \cellcolor{lightgray}$0.94$ & \cellcolor{lightgray}$0.00$ \\
\hline
Neutral & arXiv & $0.88\pm0.7$ & $0.94\pm0.04$ & $0.82$ & $0.66\pm0.09$ & $0.66\pm0.09$ & $1.00$ & $0.08$ \\
Prompt & \cellcolor{lightgray}PubMed & \cellcolor{lightgray}$0.68\pm0.17$ & \cellcolor{lightgray}$0.90\pm0.08$ & \cellcolor{lightgray}$0.83$ & \cellcolor{lightgray}$-0.10\pm0.25$ & \cellcolor{lightgray}$0.44\pm0.27$ & \cellcolor{lightgray}$0.88$ & \cellcolor{lightgray}$0.00$ \\
\hline
Prompt & arXiv & $0.87\pm0.08$ & $0.94\pm0.04$ & $0.85$ & $0.55\pm0.1$ & $0.67\pm0.08$ & $0.00$ & $0.00$ \\
Irrelevant & \cellcolor{lightgray}PubMed & \cellcolor{lightgray}$0.70\pm0.15$ & \cellcolor{lightgray}$0.89\pm0.08$ & \cellcolor{lightgray}$0.83$ & \cellcolor{lightgray}$-0.05\pm0.26$ & \cellcolor{lightgray}$0.44\pm0.27$ & \cellcolor{lightgray}$0.12$ & \cellcolor{lightgray}$0.00$ \\
\hline
\end{tabular}
\end{table}

The false positive and false negative rates in Table \ref{tab:PhysicsMarking} show the LLM was likely to state that a paper was relevant when in fact it was not. The correlations show \textit{transformer similarity} correlated better with expert review than SpaCy similarity. It was found that SpaCy similary is heavily impacted by changes in capitalisation. In some cases changes in capitalisation alone reduced the similarity to $0.3$.

\section{Discussion}
\label{sec:discussion}
In the first case study, GPT4 made less mistakes than GPT3 when extracting exact quotes from documents. This may be because of the quality of the model or GPT3's shorter maximum context length. The longest context length available for GPT3 models was 16,000 tokens which was not always sufficient to read an entire paper. When the paper was broken into multiple sections the model was less likely to find relevant quotes. We noticed that GPT3 was more likely to select quotes from the beginning of the context window than GPT4, whose quotes were more evenly distributed from the entire paper.

Across all studies it was found that transformer similarity correlates more strongly with expert opinion than than SpaCy similarity. This indicates that transformer embeddings are a better metric. This is expected given that it is able to take the positions of words into account.
The correlation between amounts of text highlighted increased with increased human judged accuracy, showing the correct trend as a measure of semantic similarity. 
However the uncertainty values were very high and more work would need to be performed to determine if this is a suitable metric. 

The highlighting tool's primary purpose is to aid a researcher in sifting through evidence and other LLM response text. 
The researchers anecdotally confirmed its value. 

We originally anticipated that SpaCy's similarity would be a better measure of semantic similarity than transformer similarity because the transformer similarity scores only ranged between 0.7 and 1, while SpaCy similarities scores had much larger ranges. Contrary to our original expectations, transformer similarity ubiquitously correlated more strongly with expert opinion. To make transformer similarity more interpretable for humans we recommend scaling these values before interpretation. This is because a transformer similarity of 0.8 is actually low, despite normal human expectations.

The second case study found that when a model was asked to provide evidence for its claims, it was slightly less accurate on all metrics including expert judgement. This is an unexpected result as normally `chain of thought' or asking a model to explain its reasoning improves performance \cite{Wei2022CoT,McGinness2024ATP}. As trustworthiness is increased when the model provides verifiable evidence for its answers, this result indicates that there is an unfortunate trade off between accuracy and trustworthiness. 
The reason could be that the model is `overloaded' when it needs to focus on multiple tasks at once. This was confirmed by the third case study (Section \ref{sec:CombiningTasks}) where model performance also decreased when the number of tasks it was asked to complete in a single call increased.

In third third case study, as expected the number of prompt tokens is approximately ten times higher for the separate condition compared to the together condition. As far as the researcher was concerned the reduced number of errors in the results was worth the the extra cost. The major issue with the together condition was the possibility for tasks to be jumbled. One area for future work is to apply more advanced parsing techniques to avoid frameshift errors and therefore make the computationally cheaper technique more desirable for researchers. Another area for future work is to provide the model with specific keywords for each call to guide the LLM responses.

The fourth case study demonstrated that SpaCy semantic similarity can be heavily impacted by the capitalisation of words for medium sized models. This is not a desirable property for a system measuring semantic similarity; we argue that writing a sentence in all capital letters makes little change to the semantic meaning. 

Readers may be tempted to think that scores such as SpaCy or transformer similarity could be used as better alternatives to subject expert review as they do not contain human biases. 
However, one needs to be careful in assuming that there is no bias when using metrics like these. There can be an illusion of objectivity, when in fact these models have been trained and validated on data which contains significant unknown biases. In this study, we highly value the opinion of experts who are part of the active research community and have observed that they demonstrate a strong awareness of their own biases.

False positive rates were consistently higher than false negative rates; GPT4 was more likely to think that an irrelevant paper was in fact relevant and most accurately screened papers when prompted to expect that papers may be irrelevant.

Overall GPT3 and GPT4 were able to find and correctly reproduce quotes from a text with 95\% and 98\% accuracy respectively. For low complexity tasks like finding the title or location of a paper, GPT4 performed with close to 100\% accuracy, but accuracy was lower for more nuanced tasks such as identifying enablers in agri-food transitions. The overall approximate average accuracy of GPT4 in answering research questions was $83\%$.

Our highlighting workflow requires \emph{keyword calibration} to determine a suitable set of keywords for each research question in an SLR.  Currently, keyword calibration is semi-automated process with an expert in the loop. The expert proposes keywords, runs highlighting experiments on sample papers and adjusts the set of keywords according to their
observations. 

From our experiments we were able to derive some guidelines for calibration. A keywords set yielding a highlighting rate of around $0.4\pm0.1$ on evidence texts often seems to be a good compromise. If it is much higher, often too many irrelevant words are highlighted. If much lower, the domain has not been covered sufficiently and relevant keywords are missing.
It is better to use evidence text than expert answers for keyword calibration as they are almost always proper English sentences. Expert assessments however can vary widely and sometimes are just lists of keywords. Hence the
proportion of highlighted words is less reliable in this case, resulting in a less meaningful hit rate. 

If the hit rate is unusually high this could be because of denser writing (less filler words) or because of denser information. The latter includes the possibility that surprising, additional insights have been unveiled. This helps to get a more complete picture of the problem. Conversely, we found that a much lower highlighting rate typically applies to irrelevant texts.

\section{Conclusion}
\label{sec:conclusion}
Large Language Models (LLMs) were used to assist interdisciplinary system scientists to conduct four Systematic Literature Reviews (SLR). The topics of the reviews were agri-food system transitions, coordinated responses to crises, sustainable transitions and automated marking. GPT-3.5 Turbo and GPT-4 Turbo had error rates of 5\% and 2\% when extracting exact quotes from research papers. Levenshtein distance accurately determined the faithfulness of quotes produced by LLMs and was robust to unexpected unicode characters and hyphenations. 

When GPT-4 Turbo completed multiple tasks in a single prompt the number of prompt tokens decreased tenfold but with significant losses in accuracy in some cases due to frameshift errors. One area for future work would be to use more advanced parsing techniques to avoid frameshift or other `jumbling' errors. 

The accuracy of the models' answers was found to decrease with complexity of the task. For very simple tasks, expert rating of LLM answer correctness was close to 100\% while for highly nuanced tasks it could be as low as 10\%. On average it was found that GPT-4 Turbo was able to extract information from papers with approximately $83\%$ accuracy. When screening papers it was found that GPT-4 Turbo was more likely to include irrelevant papers than exclude relevant papers. One area for future work is to provide the model with specific keywords to focus its answers, this may help a model focus on the desired ideas while trying to complete a nuanced task.

It was found that taking the cosine similarity of transformer embeddings of expert and LLM answers was a measure of accuracy that correlated more strongly with expert opinion than SpaCy's semantic similarity score. Nearly all of these transformer embedding cosine scores were in the range of 0.7 to 0.95. In order to make these cosine similarities more human interpretable, we recommend scaling them to take up the full range between 0 and 1. An area of future work would be to use cosine similarity of transformer embeddings for sentence-wise comparison of researcher and LLM answers in order to determine if any important pieces of information are missing.

Although highlighting is designed to assist researchers with manual checking of answers, correlation between amounts of highlighted text is showing some promise as an automated method for measuring the quality of LLM responses. Additional research would need to be conducted to see if this can be used as a valid similarity metric. Another idea for future work is to re-formulate the semantic similarity algorithm (Algorithm~\ref{alg:simlilarity}) with probabilistic logic programming. This would allow for a more flexible and expressive framework. In addition, weight parameters could be rephrased as probabilities and be learned from examples by maximum likelihood estimation.


\littleparagraph{Acknowledgements.}
This research was supported by funding from CSIRO Data61 and its Valuing Sustainability Future Science Platform initiative. We thank Enayat Moallemi for providing knowledge, expertise and data for the third case study. We thank Stephen Wan and Shima Khanehzar for helpful discussions.


%
%
%

\end{document}